\documentclass{article}

\usepackage{microtype}
\usepackage{graphicx}
\usepackage{subfigure}
\usepackage{booktabs} %

\usepackage{hyperref}
\usepackage{makecell}
\usepackage{multirow}
\usepackage{comment}

\usepackage[accepted]{icml2024}

\usepackage{amsmath}
\usepackage{amssymb}
\usepackage{mathtools}
\usepackage{amsthm}
\usepackage{pifont} 
\usepackage[capitalize,noabbrev]{cleveref}
\usepackage{arydshln} 
\usepackage{enumitem} 
\usepackage{wrapfig} 
\usepackage{hyperref} %
\usepackage{caption}

\hypersetup{
    colorlinks=true, %
    linkcolor=red, %
    filecolor=magenta, %
    urlcolor=cyan, %
}

\usepackage{xurl} %
\theoremstyle{plain}

\theoremstyle{definition}

\theoremstyle{remark}

\usepackage{xspace}
\newcommand{\xxx}[0]{\textsc{Lookahead Decoding}\xspace}

\usepackage[textsize=tiny]{todonotes}

\icmltitlerunning{Lookahead decoding}

\begin{document}

\twocolumn[
\icmltitle{Break the Sequential Dependency of LLM Inference Using \\
\xxx}


\begin{icmlauthorlist}
\icmlauthor{Yichao Fu}{ucsd}
\icmlauthor{Peter Bailis}{google}
\icmlauthor{Ion Stoica}{ucb}
\icmlauthor{Hao Zhang}{ucsd}
\end{icmlauthorlist}

\icmlaffiliation{ucsd}{UCSD}
\icmlaffiliation{google}{Google}
\icmlaffiliation{ucb}{UC Berkeley}

\icmlcorrespondingauthor{Hao Zhang}{haozhang@ucsd.edu}

\icmlkeywords{Machine Learning, ICML}

\vskip 0.3in
]

\printAffiliationsAndNotice{} %

\begin{abstract}
Autoregressive decoding of large language models (LLMs) is memory bandwidth bounded, resulting in high latency and significant wastes of the parallel processing power of modern accelerators. Existing methods for accelerating LLM decoding often require a draft model (e.g., speculative decoding), which is nontrivial to obtain and unable to generalize. In this paper, we introduce \xxx, an exact, parallel decoding algorithm that accelerates LLM decoding without needing auxiliary models or data stores. It allows trading per-step log(FLOPs) to reduce the number of total decoding steps, is more parallelizable on single or multiple modern accelerators, and is compatible with concurrent memory-efficient attention (e.g., FlashAttention). Our implementation of \xxx can speed up autoregressive decoding by up to 1.8x on MT-bench and 4x with strong scaling on multiple GPUs in code completion tasks. Our code is avialable at~\url{https://github.com/hao-ai-lab/LookaheadDecoding}
\end{abstract}

\newcommand{\vd}[0]{\mathbf{d}\xspace}
\newcommand{\vv}[0]{\mathbf{v}\xspace}
\newcommand{\vx}[0]{\mathbf{x}\xspace}
\newcommand{\vy}[0]{\mathbf{y}\xspace}
\newcommand{\vo}[0]{\mathbf{o}\xspace}
\newcommand{\xq}[0]{\mathbf{Q}\xspace}
\newcommand{\xk}[0]{\mathbf{K}\xspace}
\newcommand{\xv}[0]{\mathbf{V}\xspace}
\newcommand{\xo}[0]{\mathbf{O}\xspace}
\newcommand{\vw}[0]{\mathbf{w}\xspace}
\newcommand{\vg}[0]{\mathbf{g}\xspace}
\newcommand{\vp}[0]{\mathbf{p}\xspace}
\newcommand{\vC}[0]{\mathbf{C}\xspace}
\newcommand{\vD}[0]{\mathbf{D}\xspace}
\newcommand{\vV}[0]{\mathbf{V}\xspace}
\newcommand{\vs}[0]{\mathbf{s}\xspace}
\newcommand{\vP}[0]{\mathcal{P}\xspace}
\newcommand{\vS}[0]{\mathcal{S}\xspace}
\newcommand{\vnext}[0]{\mathbf{next}\xspace}

\section{Introduction}
\label{sec:intro}

Large language models (LLMs) are transforming the AI industry. As they are increasingly integrated into diverse applications such as search~\cite{team2023gemini} and chatbots~\cite{ouyang2022training}, generating long sequences at \emph{low-latency} using LLMs is becoming one significant requirement. However, current LLMs generate text based on~\cite{touvron2023llama,touvron2023llama2,jiang2023mistral,openai2023gpt4} \textit{autoregressive decoding}, which falls short in efficiency, primarily for two reasons. First, autoregressive decoding generates only one token at a time. Hence, the overall generation time is proportional to the number of decoding steps. Second, each decoding step largely underutilizes the parallel processing capabilities of modern accelerators (e.g., GPUs). Given the pressing need for low latency in various applications, improving autoregressive decoding remains a central challenge.

Several approaches have been proposed -- one such approach is \textit{speculative decoding}~\cite{chen2023accelerating,leviathan2023fast} and its variants~\cite{he2023rest,stern2018blockwise,medusa,EAGLE,liu2023online,miao2023specinfer}. These methods all follow a \textit{guess-and-verify} approach: they use a draft model to speculate several subsequent tokens and then use the original (base) LLM to verify these tokens in parallel. Since the draft model requires much fewer resources and the cost of verifying multiple tokens in parallel is similar to the cost of generating a single token, these methods can achieve considerable speedups. However, their speedups are bounded by the \emph{token acceptance rate} (\S\ref{sec:bspec}), i.e., the fraction of tokens generated by the draft model that passes the verification test of the base model. This is because every token that fails verification needs to be regenerated by the base model. In the worst case, if most proposed tokens fail verification, these methods may slow down the decoding process. Therefore, achieving a high acceptance rate is essential for these methods. Unfortunately, training a draft model to achieve a high acceptance rate is non-trivial, and the trained draft model does not generalize across base models and datasets.

To address these problems, this paper develops \emph{\xxx}. We build upon a key observation: autoregressive decoding can be equivalently formulated as solving a non-linear system via the fixed point Jacobi iteration method (\S\ref{sec:bg}), which we term as \emph{Jacobi decoding}~\cite{santilli-etal-2023-accelerating}. Each Jacobi decoding step can generate multiple tokens in parallel at different positions. Although these tokens may appear at incorrect positions, we can leverage this parallel generation approach to have the LLM generate several disjoint \textit{n-grams} in parallel in a single step. These n-grams could potentially be integrated into future parts of the generated sequence, pending verification by the base model to maintain the output distribution.

\xxx takes advantage of the particular characteristics of autoregressive decoding, which is bounded by the memory bandwidth--as each generated token depends on all tokens before it--rather than compute, by using the available cycles to generate and verify $n$-grams (subsequent tokens) at virtually no additional cost. In a nutshell, \xxx consists of a \textit{lookahead branch} that generates $n$-grams and a \textit{verification branch} that verifies $n$-grams, both executing in a single step. To improve efficiency, we use an $n$-gram pool to cache the historical $n$-grams generated so far. This way, \xxx can significantly reduce the latency of LLM inference just by exploiting the compute resources that autoregressive decoding would leave unused. More importantly, \xxx scales with the compute -- we show that it can linearly reduce the number of decoding steps relative to the log(FLOPs) allocated per step.

We have implemented the algorithm in both Python and CUDA, compatible with memory-efficient attention algorithms (e.g., FlashAttention~\cite{dao2023flashattention2}), and supports various sampling methods without changing the output distribution. 
We also scale it to multiple GPUs, resulting in \textit{Lookahead Parallelism}. We evaluate \xxx on the popular LLaMA-2~\cite{touvron2023llama2} models. It achieves 1.8x speedup on the challenging multi-turn chat dataset MT-Bench~\cite{zheng2023judging} and up to 4x speedup in code completion tasks with Lookahead Parallelism on 8 GPUs. \xxx showed significant potential in lowering the latency for latency-sensitive tasks.
Our contributions are summarized as follows. 

\begin{itemize}
    \item We design \xxx, a new lossless, parallel decoding algorithm to accelerate LLM inference without needing any auxiliary component. 
    \item We reveal \xxx's scaling behavior: it linearly reduces the number of decoding steps according to per-step $\log($FLOPs$)$. This enables trade-offs between the number of decoding steps and per-step FLOPs, making it future-proof.
    \item We show it benefits from the latest memory-efficient attentions and is easily parallelizable by developing its distributed CUDA implementations.
    \item  We evaluated \xxx and demonstrate its effectiveness under different settings. 
\end{itemize}

\section{Background}\label{sec:bg}

In this section, we formulate both autoregressive and Jacobi decoding from the lens of solving nonlinear systems.

\noindent \textbf{Causal Attention in Decoder Models.}\label{sec:causal}
Most contemporary LLMs are composed of two core components: token-wise modules (including MLP and normalization~\cite{ba2016layer,zhang2019root}) and attention~\cite{vaswani2023attention} modules. Tokens interact with each other in the attention modules, while in other token-wise modules, they are processed without exchanging information with each other.

The attention layer encompasses three input elements: query $\xq$, key $\xk$, and value $\xv$, with the $i$-th token in each denoted as $\xq_i$, $\xk_i$, and $\xv_i$, respectively. The attention layer executes the following operation: $\xo = \textrm{softmax}\left(\xq\xk^T\right)\xv$. A lower triangular mask applied to $\xq\xk^T$ in causal attentions (specific to decoder models) ensures that $\xo_i$ is calculated only from $\xq_i$ and $\xk_j$, $\xv_j$ where $j \leq i$.
Because all other layers in the LLM perform token-wise operations, for any given model input $\vx$ and output $\vo$, $\vo_i$ ($i$-th token in $\vo$) is exclusively influenced by $\vx_j$ ($j$-th token in $\vx$) where $j \leq i$.

\noindent \textbf{Autoregressive Decoding in LLMs.}\label{sec:ag}
LLMs take discrete integer sequences as inputs, where each integer represents a token. We notate $\vx=(x_1,x_2,...,x_s)\in\mathbb{N}^{s}$ of length $s$ as the input of the model, and $\vx_{1:m}^t=(x_1,x_2,...,x_m)$ to denote a slice of $\vx$ of length $m$ at step $t$. LLMs' output characterizes the probability distribution of the next token. The probability for the $s$-th token (i.e., the output of the $s-1$-th token) is decided by all previous input tokens, represented as $P_M(x_s|\vx_{1:s-1})$. Then, the next token input $x_s$ is obtained by sampling from $P_M(x_s|\vx_{1:s-1})$ using different methods (e.g., greedy, top-K, and top-P~\cite{JMLR:v21:19-985,holtzman2020curious}). When using greedy sampling, the next token is selected by applying an $\arg\!\max$ function on $P_M$.

We define $\vx^0$ as the prompt tokens given by the user. The LLM needs to generate an output sequence (of length $m$) from $\vx^0$. Denote $y_i$ as the token generated at step $i$. The autoregressive decoding process of $m$ tokens can be seen as solving the following $m$ problems \emph{one by one} (assume greedy sampling):
\begin{equation}\label{eq:0}
\begin{aligned}
\left\{ 
\begin{array}{l}
   y_1=\arg\!\max P_M(y_1|\vx^0)   \\
   y_2=\arg\!\max P_M(y_2|y_1,\vx^0)    \\ 
   ... \\ 
   y_m=\arg\!\max P_M(y_m|\vy_{1:m-1},\vx^0) 
\end{array}
\right.
\end{aligned}
\end{equation}

\noindent \textbf{Guess-And-Verify Paradigm.}\label{sec:bgspec}
The \textit{Guess-And-Verify} decoding paradigm speculates multiple potential future tokens and subsequently confirms the correctness of these speculations within a single decoding step. Take speculative decoding with greedy sampling as an example: at step $t$, with the prompt $\vx^0$ and tokens $\vy_{1:t-1}$ generated so far, we can use a draft model to autoregressively generate a draft sequence $\vy_{t:t+n-1}$ of length $n$. Because $\vy_{t:t+n-1}$ is known a priori, we then use the LLM to solve Eqs~\ref{eq:1} \emph{in parallel}, obtaining $\vy'_{t:t+n}$. Then, we verify if $y_{t+i}$ is equal to $y_{t+i}'$ for each $i$ from $i=0$ to $i=n-1$. If there is a match, we accept this token and proceed; otherwise, we stop checking and drop subsequent tokens. 
Finally, we update $\vy$ with all accepted tokens.
\begin{equation}\label{eq:1}
\begin{aligned}
\left\{ 
\begin{array}{l}
   y_t'=\arg\!\max P_M(y_t|\vy_{1:t-1},\vx^0)   \\
   y_{t+1}'=\arg\!\max P_M(y_{t+1}|\vy_{1:t},\vx^0) \\     
   ...  \\
   y_{t+n}'=\arg\!\max P_M(y_{t+n}|\vy_{1:t+n-1},\vx^0) 
\end{array}
\right.
\end{aligned}
\end{equation}
As stated in \S\ref{sec:intro}, these approaches depend on a good draft model, which is hard to obtain and cannot generalize.

\noindent \textbf{Jacobi Decoding.}
By notating  $f(y_i,\vy_{1:i-1},\vx^0)=y_i-\arg\!\max P_M(y_i|\vy_{1:i-1},\vx^0)$, we can transform Eqs~\ref{eq:0} into the following non-linear system of equations~\cite{song2021accelerating,santilli-etal-2023-accelerating}:
\begin{equation}\label{eq:2}
\begin{aligned}
\left\{ 
\begin{array}{l}
   f(y_1,\vx^0)=0 \\
   f(y_2, y_1,\vx^0)=0 \\     
   ... \\
   f(y_m,\vy_{1:m-1},\vx^0)=0 
\end{array}
\right.
\end{aligned}
\end{equation}

We can solve this non-linear system using Jacobi iteration by iteratively updating all $y_i$ from a random initial guess $\vy^0$, along the \emph{trajectory} $\vy^1, ..., \vy^t, ...$, until converging to the fixed point solution $\vy^m$. We detail this algorithm, termed as \emph{Jacobi decoding}, in Appendix Algorithm \ref{alg:jacobi}. This process guarantees to return the solution of all $m$ variables $y_i$ in \emph{at most }$m$ iterations, as the very first token of each Jacobi update matches autoregressive decoding. Sometimes, more than one token might be correctly generated in a single iteration, potentially reducing the number of decoding steps. It is worth noting that, as $\vy^t$ is generated based on the past value $\vy^{t-1}$ on the trajectory, any two adjacent tokens from $\vy^{t-1}$ and $\vy^{t}$ can form a meaningful 2-gram.

\noindent \textbf{Limitations of Jacobi Decoding.} Empirically, we observe Jacobi decoding can hardly reduce decoding steps, even if it can generate multiple tokens per step. This is because the generated tokens are often put in the wrong positions of the sequence, and correctly placed tokens are frequently replaced by subsequent Jacobi iterations. These prevent it from achieving wall-clock speedup.

\section{\xxx}\label{sec:detail}

\begin{figure}
\begin{center}
\caption{Workflow of \xxx with $W=5$, $N=3$, and $G=2$. For each decoding step, we do the following. (1) Generate one token at each position in the lookahead branch; (2) Verify and accept 3-grams (searched from the 3-gram pool) with the verification branch; (3) Collect and cache newly generated 3-grams in the pool from lookahead branch trajectories. (4) Update the lookahead branch to maintain a fixed window size.
}
\vskip -0.1in
\centerline{\includegraphics[width=1.0\columnwidth]{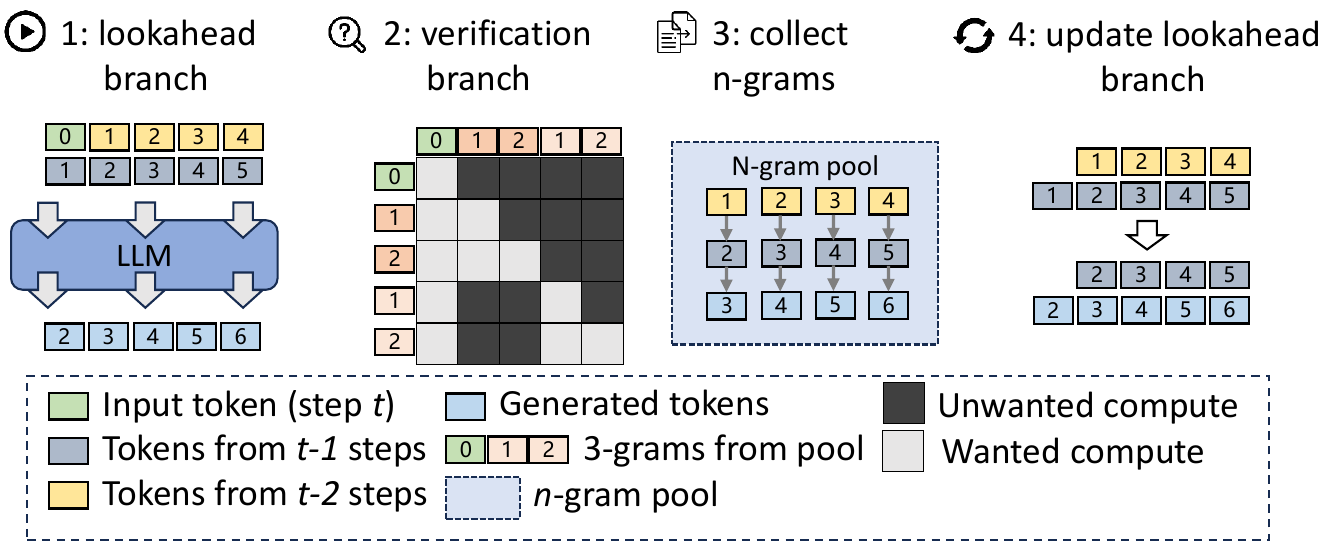}}
\label{fig:ppt}
\end{center}
\vskip -0.5in
\end{figure}

\xxx leverages Jacobi decoding's ability to generate many tokens in one step but addresses its limitation. Fig.~\ref{fig:ppt} illustrates its workflow.
The key design in \xxx is to keep track of the trajectory of Jacobi decoding and generate $n$-gram from this trajectory. This is achieved by maintaining a fixed-sized 2D window, with the two dimensions corresponding to the sequence and the time axis, respectively, to generate multiple disjoint $n$-grams from the Jacobi iteration trajectory in parallel. We call this process the \emph{lookahead branch}.
In addition, \xxx introduces an $n$-gram pool to cache these $n$-grams generated along the trajectory.
Promising $n$-gram candidates are verified later by a designed \emph{verification branch} to preserve the LLM's output distribution; if passing verification, those disjoint n-grams are integrated into the sequence.
The detailed algorithm is shown in Algorithm~\ref{alg:lookahead} in Appendix.

\subsection{Lookahead Branch}
\xxx uses a fixed-sized 2D window for efficient $n$-gram generation. In contrast to the original Jacobi decoding, which only uses the history tokens from the last step (or equivalently, it generates 2-grams), \xxx generates many $n$-grams, with $n \ge 2$, in parallel by using the $n-1$ past steps' history tokens, effectively leveraging more information from the trajectory.
The fixed-sized 2D window in the lookahead branch is characterized by two parameters: (1) $W$ defines the \emph{lookahead} size into future token positions to conduct parallel decoding; (2) $N$ defines the \emph{lookback} steps into the past Jacobi trajectory to retrieve $n$-grams. See Algorithm~\ref{alg:lookahead} for a detailed process.

An example of the lookahead branch with $W=5$ and $N=4$ is in Fig.~\ref{fig:att} (b), in which we look back $N-1=3$ steps and look ahead $5$ tokens for each step. The blue token with the digit 0 is the current step's ($t$) input, and the orange, green, and red tokens were generated in previous lookahead branches at steps $t-3$, $t-2$, and $t-1$, respectively. The digit on each token shows its relative position to the current input (i.e., the blue one labeled as 0). In the present stage, we perform a modified Jacobi iteration to generate new tokens for all 5 positions, following the trajectory formed by the preceding 3 steps. Once generated, we collect and cache them in the $n$-gram pool ($n=4$) -- for instance, a 4-gram consists of the orange token at position 1, the green token at position 2, the red token at position 3, and a newly generated token.

The most outdated tokens in both dimensions (time and sequence) will be removed, and newly generated tokens will be appended to the lookahead branch to maintain a fixed window size for each step. For example, we will remove all orange and green tokens with position 1 in Fig.~\ref{fig:att}. We then form a new lookahead branch with green tokens with indices 2, 3, 4, 5, all red tokens, and all newly generated tokens for the next step.

\subsection{Verification Branch}\label{sec:verification}
\xxx preserves the output distribution via its verification branch. 
We first discuss how to verify in greedy sampling. 
Recall in speculative decoding: the verification is performed by sending the draft tokens to the LLM to get an output for each draft token, then progressively checking if the last token's corresponding output, generated by the target LLM, exactly matches the draft token itself (\S\ref{sec:bgspec}). The verification branch in \xxx resembles this process, despite verifying \emph{many draft $n$-gram} candidates in parallel. In particular, We first look up from the $n$-gram pool for ``promising'' $n$-grams -- by checking if a $n$-gram starts with a token that exactly matches the last token of the current ongoing sequence. We then use the LLM to verify all these $n$-grams in parallel, following a similar fashion as in speculative decoding. See Algorithm~\ref{alg:greedy} in the Appendix for the detailed procedures.

We next discuss how to support more advanced sampling. Previous research~\cite{miao2023specinfer} has developed efficient tree-based verification for speculative decoding with sampling support, where multiple draft sequences derived from a token tree can be verified in parallel. However, it does not apply to \xxx as our verification works on disjoint $n$-grams instead of trees. We improve it by progressively verifying along the $n$-gram length and removing $n$-grams with mismatched prefixes. Besides, speculative decoding style verification requires the probability distribution where the \emph{draft token} is sampled to update the probability distribution when the draft token is rejected. Because we store all $n$-grams in a pool instead of discarding them each step,  we would need huge memory to store the probability distributions (each of vocabulary size) for the entire $n$-gram pool. The key to overcome this is to leverage the mechanism that the verification is indifferent to how draft tokens were sampled -- different sampling methods (e.g., greedy sampling) only influence the acceptance rate but keep the output distribution. We can force greedy sampling at the $n$-gram generation (lookahead branch), in which the probability distribution degenerates into a one-hot vector. Hence we only need to store which token is selected. We elaborate the approach in Algorithm~\ref{alg:sample}, prove its correctness in Appendix~\ref{app:prof-sample}, and verify its quality and speedups in~\S\ref{sec:quality}.

It is expected to have an increasingly large $n$-gram cache hence a growing verification branch as decoding progresses. We set a cap of $G$ to limit the maximum number of promising candidates run in parallel in the verification branch to manage the verification cost. Empirically we suggest to set $G$ proportional to $W$ to balance generation and verification. In practice, we simply set $G=W$.

\subsection{Decode, Predict, and Verify in The Same Step}

\begin{figure}
\begin{center}
\centerline{\includegraphics[width=1.0\columnwidth]{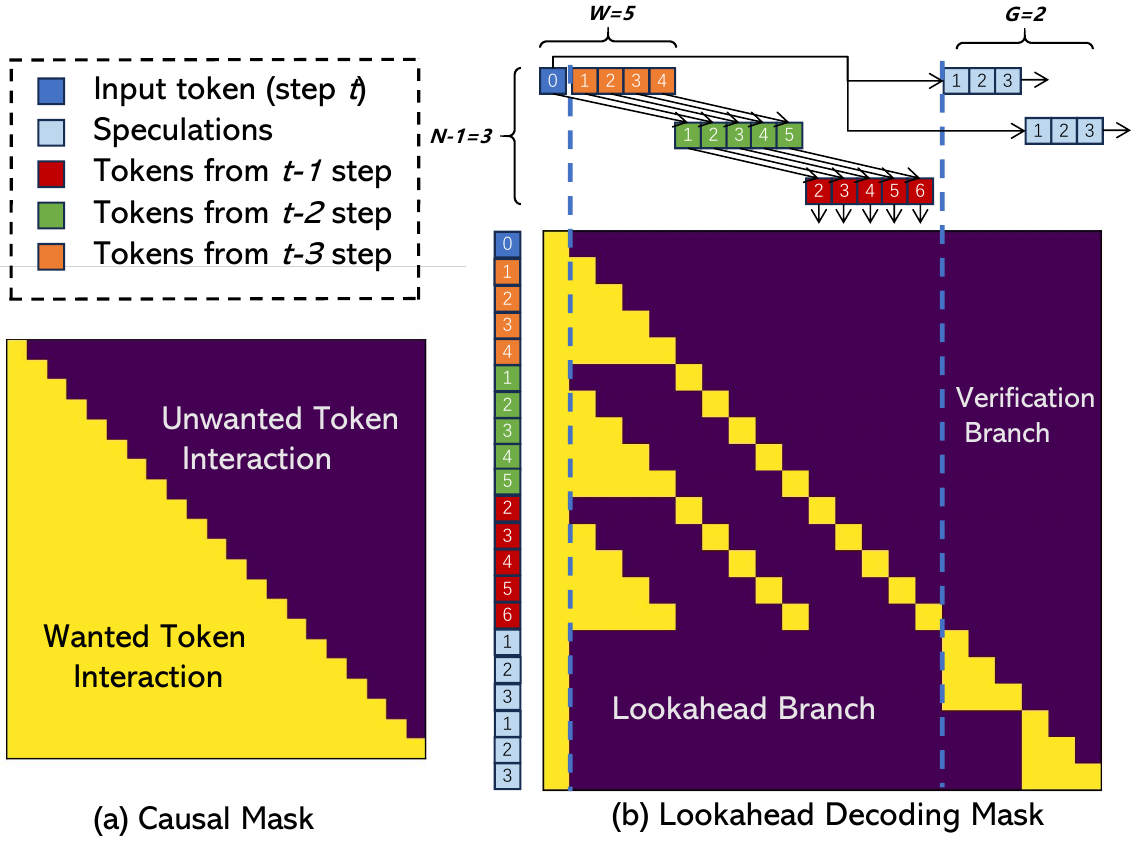}}
\caption{(a) Causal mask for decoder models. (b) Attention mask for \xxx with $W=5$, $N=4$, and $G=2$. Digits on tokens indicate relative positions. }
\label{fig:att}
\end{center}
\vskip -0.4in
\end{figure}

At execution, the lookahead and verification branches can be integrated into one decoding step to leverage parallel processing. This requires a designated attention mask, as shown in Fig.~\ref{fig:att} (b). This attention mask is straightforwardly derived following the principle that each token is only visible to the tokens with a larger position index than itself (\S\ref{sec:causal}). For example, only the green token at position 5 and all orange tokens are visible to the red token 6. The tokens in the lookahead branch are not visible to the tokens in the verification branch, and vice versa.  

\noindent \textbf{Integration with FlashAttention.}
FlashAttention~\cite{dao2022flashattention,dao2023flashattention2} can vastly accelerate the training and inference of LLMs by saving memory I/O on the slow memory hierarchy. It forces a causal mask (e.g., Fig.~\ref{fig:att} (a)) to avoid all token interactions outside a lower triangular scope, which is not suitable for \xxx as we take a more subtle attention mask (e.g., Fig.~\ref{fig:att} (b)) for different $W$, $N$, and $G$. To solve this, we hardcode \xxx's attention pattern with adjustable $W$, $N$, and $G$ in FlashAttention. Applying FlashAttention to \xxx brings about 20\% end-to-end speedup compared to a straightforward implementation on top of native PyTorch in our experiments (\S\ref{sec:flash}).

\subsection{Lookahead Parallelism}\label{sec:dist}
\xxx is easy to parallelize on multiple GPUs for both lookahead and verification branches. 
Parallelizing the lookahead branch is achieved by noting that the lookahead computation is composed of several disjoint branches. For example, the branch with green 1 and red 2 tokens does not have interaction with the branch with the tokens green 3 and red 4 in Fig.~\ref{fig:att} (b). We can put these disjoint branches onto different GPUs without introducing communication during the inference computation. 
Parallelizing the verification branch is done by assigning multiple $n$-gram candidates to different devices. Because the verification of each candidate, by design, is independent of others, this will not cause communication. 

\begin{wrapfigure}{r}{0.45\linewidth}
    \centering
    \vspace{-1.5em}
    \includegraphics[width=\linewidth]{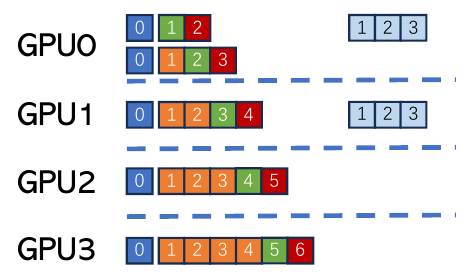}
    \vspace{-2.0em}
    \caption{Distribute the workload of the lookahead branch and verification branch in Fig~\ref{fig:att} (b) to 4 GPUs with lookahead parallelism, which can avoid communication during the forward pass.}
    \label{fig:dist}
    \vspace{-1.0em}
\end{wrapfigure}

Fig.~\ref{fig:dist} shows an example of parallelizing the lookahead branch and verification branch in Fig.~\ref{fig:att} (b) to four GPUs. This workload allocation will have the orange token 0,1,2,3 and the input token 0 be redundantly placed and computed. However, it can essentially save communication volume during the whole forward pass. We only need to synchronize the generated tokens on each device after the forward pass. We can further scale the $W$, $N$, and $G$ with multiple GPUs' increased FLOPs to obtain a lower latency according to \xxx's scalability (\S\ref{sec:scale}).  

We name this new parallelism as \textit{lookahead parallelism} (LP). Unlike previous parallelism methods (including pipeline and tensor parallelisms) that shard the model parameters or states across different GPUs, LP maintains an entire copy of the model for each GPU (thus needing more memory) and allows distributing tokens to different GPUs.
Hence, LP is advantageous in inference as it introduces near-zero communication per step while existing model parallelism methods~\cite{narayanan2021efficient, shoeybi2019megatron} involve a large communication overhead on the critical path of each decoding step.

\section{Scaling Law of \xxx}
\label{sec:scale}
Since \xxx introduces flexible parameters $W$ and $N$ associated with the cost of each parallel decoding step. This section investigates the scaling law between compute FLOPs and the theoretical speedup, and compares it to speculative decoding.

\subsection{Estimating Speedup for Speculative Decoding}\label{sec:bspec}

Speculative decoding uses the draft model to speculate one token sequence at each step. We represent the probability of each token in the sequence passing the verification of the LLM by $\beta$ (acceptance rate) and notate its expectation $E(\beta) = \alpha$. If we use the draft model to guess $\gamma$ tokens per step, the expectation of the number of accepted tokens is denoted as~\cite{leviathan2023fast}:
\begin{eqnarray}
\footnotesize
\label{for:spec}
 E(\#tokens)  = \frac{1-\alpha^{\gamma + 1}}{1-\alpha}.
\end{eqnarray}

Instead of speculating one sequence every time, we would speculate $b$ sequences. We assume that $b$ sequences, each of $\gamma$ tokens, are sampled as each token will have the same acceptance rate of $\beta$. Under this setting, the expectation of the number of accepted tokens is denoted as follows: 
\begin{eqnarray}
\footnotesize
\label{for:bspec}
E(\#tokens) = (\gamma+1)-\sum\limits_{i=1}^{\gamma}(1-\alpha^i)^b.   
\end{eqnarray}
See derivations in Appendix~\ref{app:exp} for Eq.~\ref{for:spec} and Eq.~\ref{for:bspec}. Note that when $b=1$, Eq.~\ref{for:bspec} falls back to Eq.~\ref{for:spec}.

\subsection{Estimating Speedup for \xxx }

We define the $\vS= \textit{step compression ratio}$ as the number of autoregressive steps divided by the number of \xxx steps to generate the same length of the sequence. As the number of generated tokens equals the autoregressive steps, it can be denoted as:
\begin{eqnarray}\label{eq:cp}
\vS = \frac{\#generated\ tokens}{\#\xxx\ steps}
\end{eqnarray}

\begin{figure}[ht]
\begin{center}
\centerline{\includegraphics[width=\columnwidth]{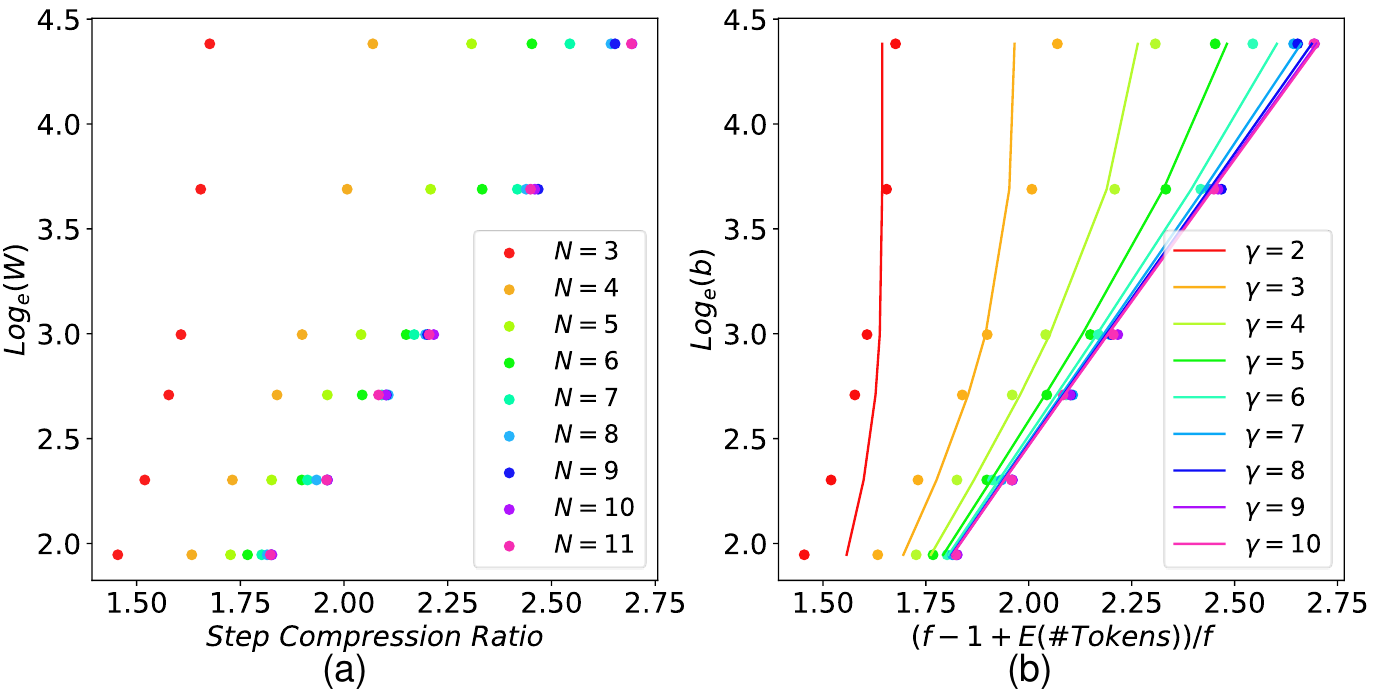}}
\caption{(a) Relation of $W,N,G$ and $\vS$ for LLaMA-2-Chat-7B on MT-Bench. (b) When we assume a setting with $\alpha=0.425$ and $f=3.106$, the trend of our formulation.}  
\label{fig:scaling}
\end{center}
\vskip -0.4in
\end{figure}

\xxx speculates $b$ sequences every time as in Eq.~\ref{for:bspec}. In each step, we will search $n$-grams in the pool starting with the current input token and have at most $G$ speculations of length $N-1$. As we set $G=W$ (\S\ref{sec:verification}), we have $G=W=b$ and $N-1=\gamma$ using the notations in Eq.~\ref{for:bspec}. In practice,  we cannot expect each step to have equally good speculations (i.e., acceptance rate with $E(\beta)=\alpha$). We assume that, on average, for every $f$ step, we have one good speculation with $E(\#tokens)$ tokens accepted, and for the other $f-1$ steps, we fall back to autoregressive decoding due to bad speculations. We use this $f$ to bridge $\vS$ and $E(\#tokens)$ per step as follows: 
\begin{eqnarray}
\vS = (f-1 + E(\#tokens)) / f.
\end{eqnarray}
We can plot the curve indicated by our formulation with one specific setting as in Fig.~\ref{fig:scaling} (b). We find that the trend of our empirical experiments (LLaMA-2-Chat-7B on MT-Bench with $G=W$ as in Fig.~\ref{fig:scaling} (a)) align well with the formulation to some extent. From this formulation, we conclude that we can linearly reduce the number of decoding steps according to per-step $\log(b)$ given a large enough $\gamma$. In contrast to speculative decoding, \xxx will not meet an upper bound indicated in Eq.~\ref{for:spec} by simultaneously increasing $\gamma$ and $b$. This reveals the \emph{scaling law} of \xxx to linearly reduce decoding steps according to per-step $\log($FLOPs$)$ given a large enough $N$, since per-step FLOPs is roughly proportional to the number of input tokens (i.e., $(W+G)*(N-1)$). The scaling law also suggests \xxx's strong scaling to multiple GPUs, in which we can obtain an even greater per-token latency reduction by using more FLOPs, which is advantageous for latency-sensitive tasks. 

\section{Evaluation Results}\label{sec:eval}

\textbf{Model and testbed.} We used various versions of the LLaMA-2~\cite{touvron2023llama2} and CodeLlama~\cite{roziere2023code} models, including the 7B, 13B, 34B, and 70B sizes, 
on two GPU setups \textit{S1} and \textit{S2}. \textit{S1} is equipped with NVIDIA A100 GPUs with 80GB of memory. On \textit{S1}, the 7B, 13B, and 34B models are deployed on a single A100, while the 70B model utilizes 2 A100s with pipeline parallelism supported by Accelerate~\cite{accelerate}. \textit{S2} is a DGX machine with 8 NVIDIA A100 GPUs with 40GB memory and NVLink. All models serve with FP16 precision and batch of 1 if not specified~\cite{medusa,he2023rest}.

\textbf{Datasets.} We benchmarked \xxx's performance across a broad spectrum of datasets and tasks. MT-Bench~\cite{zheng2023judging} is a diverse set of multi-turn questions with many \emph{unique tokens}. GSM8K~\cite{cobbe2021gsm8k} contains a set of math questions, in which we use the first 1k questions. HumanEval~\cite{chen2021evaluating} covers both code completion and infilling tasks. We also test on MBPP~\cite{austin2021program} dataset for instruction-based code generation, and on ClassEval~\cite{du2023classeval} for class-level code completion. To control generation length in code generation tasks, we set the maximum sequence length to 512 and 2,048 on HumanEval and ClassEval, respectively, aligned with prior setups~\cite{bigcode-evaluation-harness,du2023classeval}. Tab.~\ref{tab:model-1} lists detailed settings. In addition, we validate the effectiveness of sampling (\S\ref{sec:verification}) on 
XSum~\cite{xsum-emnlp} and 
CNN/Daily Mail~\cite{see2017get} datasets.

\textbf{Baseline Settings.}
Our primary baseline is HuggingFace's implementation of greedy search~\cite{wolf-etal-2020-transformers}. Additionally, we employ FlashAttention~\cite{dao2022flashattention,dao2023flashattention2} as a stronger baseline to assess the performance of FlashAttention empowered \xxx. In distributed settings, we evaluate LP against TP (supported by deepspeed~\cite{aminabadi2022deepspeed}) and PP (supported by accelerate~\cite{accelerate}). We measure the throughput of single batch inference against these baseline settings~\cite{medusa,he2023rest}.

\begin{table}
\caption{Experimental settings for \S\ref{sec:bench} and \S\ref{sec:flash}.} 
\vskip -0.15in
\setlength\tabcolsep{2pt}
\label{tab:model-1}
\begin{center}
\begin{scriptsize}
\begin{sc}
\begin{tabular}{ccccc}
\toprule
Server & Parallel. & Model & Model Size & Dataset  \\
\midrule
\multirow{3}*{\textit{S1}}&\multirow{3}*{w/o LP}&LLaMA-2-chat    & 7B, 13B, 70B & MT-Bench  \\
~&~&CodeLLaMA    & 7B, 13B, 34B & HumanEval   \\
~&~&CodeLLaMA-Inst   & 7B, 13B, 34B & MBPP, GSM8K   \\
\midrule
\multirow{3}*{\textit{S2}}&\multirow{3}*{w/ LP} &LLaMA-2-chat    & 7B, 13B & MT-Bench  \\
~&~&CodeLLaMA    & 7B, 13B & HumanEval   \\
~&~&CodeLLaMA-Python   & 7B, 13B & ClassEval   \\
\bottomrule
\end{tabular}
\end{sc}
\end{scriptsize}
\end{center}
 \vskip -0.4in
\end{table}

\begin{figure*}
\begin{center}
			\includegraphics[width=1.0\textwidth]{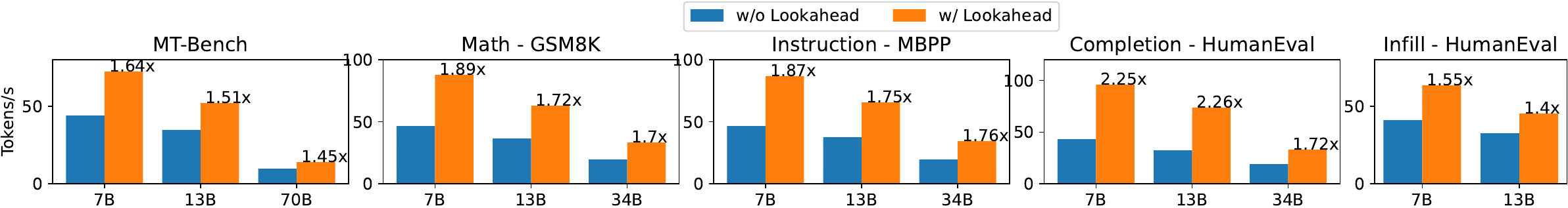} 
\vskip -0.1in
\caption{Throughput of \xxx on various dataset without FlashAttention and distributed serving}
\label{fig:throughput}
\end{center}
\vskip -0.2in
\end{figure*}

\begin{figure}
\begin{center}
			\includegraphics[width=1.0\columnwidth]{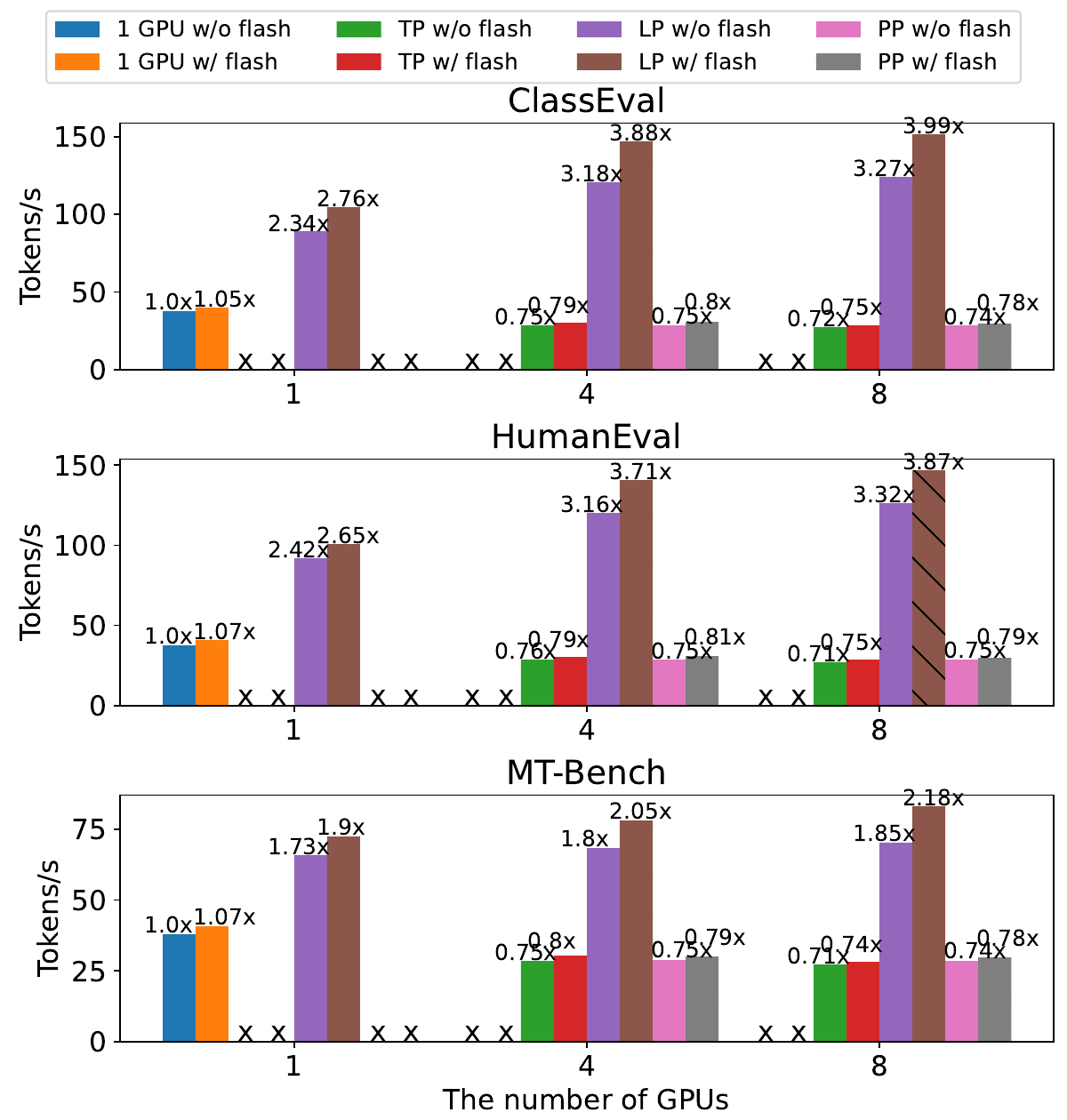} 
\vskip -0.1in
\caption{Throughput of \xxx with multiple GPUs and FlashAttention for 7B models}
\label{fig:throughput-7b}
\end{center}
\vskip -0.2in
\end{figure}

\begin{figure}
\begin{center}
			\includegraphics[width=1.0\columnwidth]{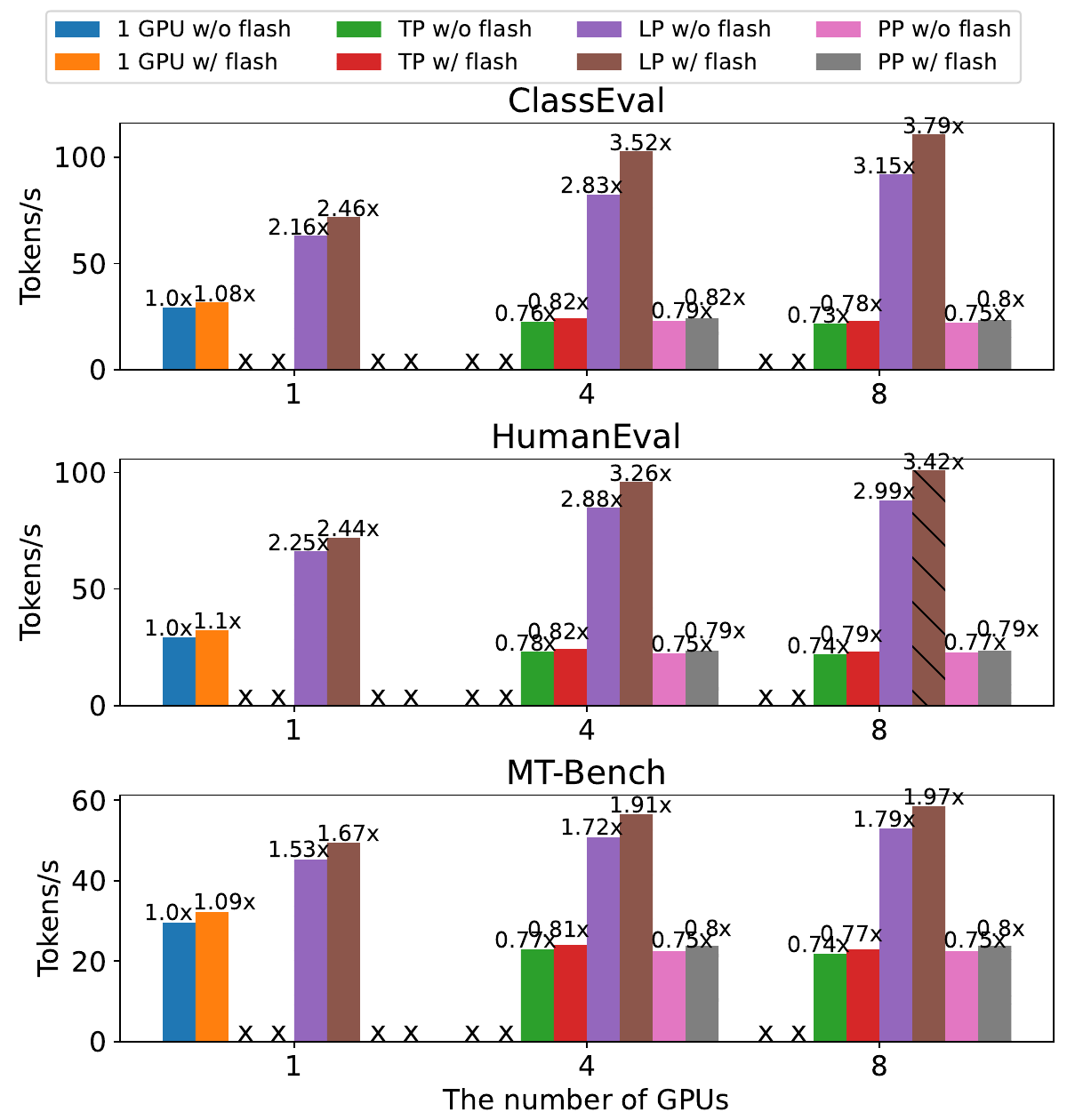} 
\vskip -0.1in
\caption{Throughput of \xxx with multiple GPUs and FlashAttention for 13B models}
\label{fig:throughput-13b}
\end{center}
\vskip -0.2in
\end{figure}

\subsection{End-to-end Performance}\label{sec:bench}

Fig.~\ref{fig:throughput} shows the end-to-end performance of \xxx when compared with HuggingFace's implementation of greedy search on \textit{S1}. 
The used tasks and models are shown in Tab.~\ref{tab:model-1}.
Across various datasets, \xxx demonstrates a 1.5x-2.3x speedup. Generally, our method exhibits better performance in code completion tasks (e.g., 2.3x), given the higher occurrence of repetitive tokens during code completions, making predictions easier. Besides, smaller models also exhibit a higher speedup when compared to larger models. This is because \xxx trades per-step FLOPs with a step compression ratio (\S\ref{sec:scale}). A larger model requires more FLOPs and quickly hits the GPU FLOPs cap compared to a smaller model. So, it shows a lower ability to compress decoding steps given the same GPU setting.

\subsection{Performance with LP and FlashAttention}\label{sec:flash}

We evaluated the performance of \xxx with LP and FlashAttention augmentation on \textit{S2} with greedy search. The used tasks and models are shown in Tab.~\ref{tab:model-1}. The results for the 7B and 13B models are in Fig.~\ref{fig:throughput-7b} and Fig.~\ref{fig:throughput-13b}, respectively. FlashAttention speeds up the PyTorch implementation of \xxx by 20\%. Notably, FlashAttention-integrated \xxx shows 1.8x speedups for the 7B model on MT-Bench compared with autoregressive decoding with FlashAttention (i.e., 1.9x vs 1.07x in Fig.~\ref{fig:throughput-7b}). We did a strong scaling of the workloads to multiple GPUs for distributed settings (i.e., increasing GPUs but not increasing workloads). The multiple GPU settings of both TP (w/ DeepSpeed) and PP (w/ Accelerate) bring slowdowns (i.e., 0.75x-0.82x). The results echos DeepSpeed's documentation~\cite{deepspeed2023tp}. However, with \xxx, we can further utilize the FLOPs of multiple GPUs to reduce the inference latency (e.g., 4x on ClassEval).

\begin{table}[t]
\caption{Sampling with \xxx on CNN/Daily Mail and XSum. A temperature (Temp.) of 0.0 equals greedy search. ``AR.'' is autoregressive and ``LA.'' is \xxx. Rouge scores, speedups against autoregressive, and compression ratio ($\vS$) are reported.}
\setlength\tabcolsep{1pt}
\vskip -0.15in
\label{tab:sample-perf}
\begin{center}
\begin{scriptsize}
\begin{sc}
\begin{tabular}{cccccccc}
\toprule
 Dataset & Temp. & Method & Rouge-1 & Rouge-2 & Rouge-L  & Speedups &  $\vS$ \\
\toprule %
\multirow{4}{*}{CNN.}& \multirow{2}{*}{1.0} & AR. &  36.55 & 13.20 & 22.68     &1.00x  & 1.00x\\ %
&& LA.& 36.53 & 13.27 & 22.71   & 1.46x & 1.64x\\ %
&\multirow{2}{*}{0.0} & AR.& 37.79 & 14.59 & 23.96      & 1.00x & 1.00x \\ %
&&LA.& 37.79 & 14.59 & 23.96     &  1.57x & 1.72x\\ %
\midrule[0.8pt]
\multirow{4}{*}{Xsum} & \multirow{2}{*}{1.0} & AR.& 19.15 & 4.53 & 12.84      & 1.00x & 1.00x \\ %
& &LA.&  19.20 & 4.53 & 12.87    & 1.50x & 1.67x \\ %
& \multirow{2}{*}{0.0} & AR.&  19.38 & 4.78  & 13.05 &     1.00x & 1.00x \\ %
& &LA.&   19.39 & 4.79  & 13.06  & 1.60x & 1.77x \\ %
\bottomrule
\end{tabular}
\end{sc}
\end{scriptsize}
\end{center}
\end{table}

\begin{table}
\caption{Compare the effectiveness of both lookahead and verification branch on MT-Bench on A100. FlashAttention is activated. We show the speedups against autoregressive decoding and the compression ratio ($\vS$). }
\vskip -0.15in
\setlength\tabcolsep{3pt}
\label{tab:prompt}
\begin{center}
\begin{scriptsize}
\begin{sc}
\begin{tabular}{ccccc}
\toprule
Tag &Setting (N, W, G) & \textit{Prompt as Ref.} & Speedups	& $\vS$  \\ %
\midrule
\ding{172} &Autoregressive & \ding{55}  & 1.00x & 1.00 \\ %
\ding{173} &Prompt Lookup  & \ding{51}  & 1.44x & 1.55 \\ %
\ding{174} &(10, 1, 3) &  \ding{51}& 1.36x & 1.45 \\ %
\ding{175} &(5, 1, 10) &  \ding{51}& 1.36x & 1.51 \\ %
\ding{176} &(5, 1, 30) &  \ding{55}  & 1.04x & 1.12 \\ %
\ding{177} &(5, 1, 30) &  \ding{51}  & 1.46x & 1.59 \\ %
\ding{178} &(5, 30, 1)& \ding{55} & 1.61x & 1.79 \\ %
\ding{179} &(5, 15, 15) & \ding{55} & 1.78x & 1.96 \\ %
\ding{180} &(5, 15, 15) & \ding{51}& \textbf{1.88x} & \textbf{2.05}\\ %
\bottomrule
\end{tabular}
\end{sc}
\end{scriptsize}
\end{center}
\vskip -0.2in
\end{table}

\subsection{Generation Quality of \xxx}\label{sec:quality}
We assess the generation quality of \xxx on LLaMA-2-7B-Chat model with the prompts in Appendix~\ref{app:sum} on summarization datasets~\cite{chen2023accelerating,leviathan2023fast} in Tab.~\ref{tab:sample-perf}. Whether the sampling is activated, \xxx can reserve the output distribution quality, which is evaluated in rouge-1, rouge-2, and rouge-L~\cite{lin-2004-rouge}, while achieving 1.46x-1.60x speedups compared with autoregressive decoding. Using sampling gives smaller speedups as the acceptance ratio is lower according to the sampling verification algorithm~\ref{alg:sample}, which aligns with the results in the previous research~\cite{chen2023accelerating,leviathan2023fast}. We further verify that using greedy sampling and advanced integrations will not change the generation quality in Appendix~\ref{app:verify}.

\subsection{Ablation Study}\label{sec:ablation}
In this section, we study the importance of the lookahead and verification branch in achieving a high speedup. We experiment on LLaMA-2-7B-Chat and MT-Bench on \textit{S1} with various settings. The results are shown in Tab.~\ref{tab:prompt}. 

We ablate the \emph{importance of lookahead branch} by comparing the performance of using a lookahead branch to the recent methods of using \textit{prompts as reference}~\cite{yang2023inference,saxena2023prompt}. This comparison assumes that  \xxx does not use the prompt to build the n-gram pool. We use the implementation in transformers v4.37 of prompt lookup as a baseline (\ding{173}, with prompt\_lookup\_num\_tokens=10). We also use prompt to build n-gram pool to augment \xxx  (\ding{174}\ding{175}\ding{177}\ding{180}). The results show that although using a minimal lookahead branch ($W=1$) with various $N, G$ settings (\ding{174}\ding{175}\ding{176}\ding{177}) can obtain a decent speedup on MT-Bench, it is still not as good as using balanced branches (\ding{179}). We can find that prompt lookup can surpass \textit{prompt as reference} implementation in \xxx. This is because our method checks if $n$-gram starts with \emph{one} token that exactly matches the last generated token while prompt lookup in transformers v4.37 checks \emph{several} starting tokens for a better speculation.

We ablate the \emph{importance of verification branch} by reporting the speedup of using a tiny verification branch and a large lookahead branch (\ding{178}, $G=1$) . It shows lower performance due to lower potential in accepting speculations compared with a balanced branches (\ding{179}).  

Besides, our evaluation shows that using \textit{prompt as reference} can further boost \xxx (\ding{179} and \ding{180}). We have integrated them in our implementation.

\subsection{Discussion and Limitation}

\begin{wraptable}{r}{0.5\linewidth}
\setlength\tabcolsep{2pt}
\vspace{-0.15in}
\caption{Good Config. of \xxx on A100 GPUs with $G=W$.}
\label{tab:config}
\vspace{-0.2in}
\begin{center}
\begin{scriptsize}
\begin{sc}
\resizebox{\linewidth}{!}{
\begin{tabular}{ccc}
\toprule
Model &	Window Size ($W$)	& N-gram Size ($N$) \\
\midrule
7B & 15 & 5 \\
13B & 10 & 5 \\ 
34B & 7 & 5 \\
\bottomrule
\end{tabular}
}
\end{sc}
\end{scriptsize}
\end{center}
\vspace{-0.2in}
\end{wraptable}

The main limitation of \xxx is that it requires extra computations. Our experimental results show that on A100, the configuration in Tab.~\ref{tab:config} works near optimally in most cases for single batch serving. Because the per-step FLOPs are roughly proportional to the number of per-step input tokens, which is $(W+G)*(N-1)$. If we ignore the attention cost's increase with sequence length, the 7B, 13B, and 34B models require 120x, 80x, and 56x extra FLOPs per step, respectively. Since the LLM decoding is memory bandwidth-bound rather than compute-bound, these extra FLOPs only turn into a limited wall-clock slowdown for each step.

\begin{figure}
\begin{center}
\centerline{\includegraphics[width=1.0\columnwidth]{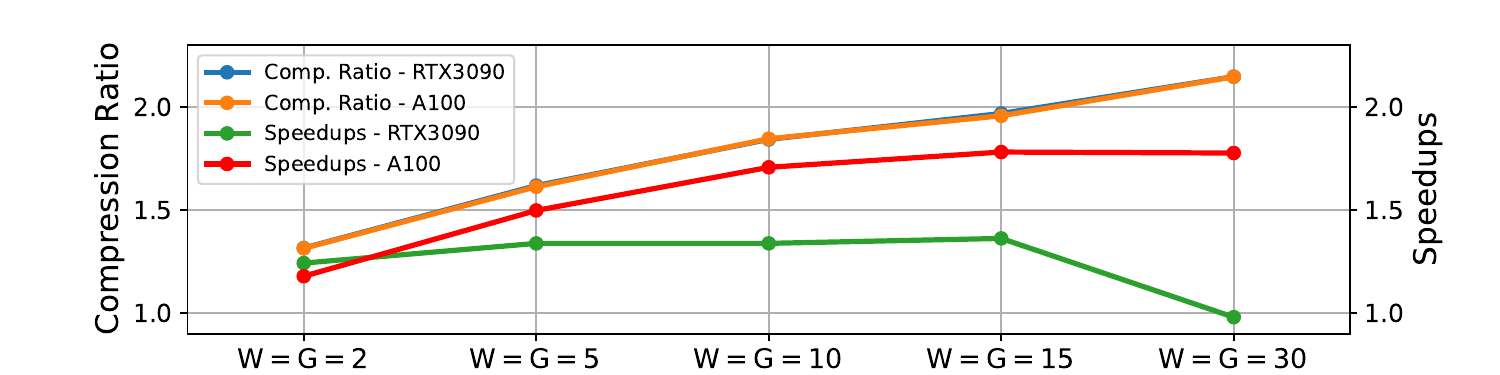}}
\vskip -0.1in
\caption{Compression ratio($\vS$) and speedups of \xxx on RTX 3090 and A100 with $N=5$, all with FlashAttention. The blue and orange curves of $\vS$ overlap as the device does not affect the ratio.}

\label{fig:overhead}
\end{center}
\vskip -0.4in
\end{figure}

Given this, \xxx needs large surplus FLOPs to obtain high speedups. Running in compute-bound environments (e.g., serving with a large batch size) may cause slowdowns. Another example is shown in Fig.~\ref{fig:overhead}, where lower speedup is observed when the GPU's cap FLOPs is smaller (e.g., on RTX 3090 GPUs).

Based on \S\ref{sec:scale}, we need to exponentially increase the per-step FLOPs to obtain a linear reduction in decoding steps. Hence, the setting in Tab.~\ref{tab:config} faces a diminishing return. However, when FLOPs are not rich, we see that a gentle speedup (e.g., $30\%$ on RTX 3090 and $> 50\%$ on A100) on MT-Bench easily achievable, as in Fig.~\ref{fig:overhead}, which is a free lunch that requires no extra model, training, or changing the output distribution. 

\section{Related Work}
 Speculative decoding~\cite{chen2023accelerating,leviathan2023fast} pioneer in speedup autoregressive decoding with a draft model. Different methods for obtaining speculations are researched. Specinfer~\cite{miao2023specinfer} uses many draft models obtained from distillation, quantization, and
pruning to conduct speculations together. Medusa~\cite{medusa}, OSD~\cite{liu2023online}, and EAGLE~\cite{EAGLE} use training to obtain a draft model. REST~\cite{he2023rest} uses the finetuning dataset itself as a datastore to lookup speculations, while other works~\cite{yang2023inference,saxena2023prompt} uses prompt as a reference for speculations. Different from these methods, \xxx uses LLM's parallel generation ability for speculations. Sampling methods are also researched. Specinfer maintains output distribution by a tree-based sampling algorithm. Medusa uses a typical acceptance scheme to accelerate when the temperature is large but does not persist on an exact output distribution. \xxx follows Specinfer to maintain output distribution but with multiple disjoint $n$-grams.

\section{Conclusion}\label{sec:conclusion}
In this paper, we present \xxx to parallelize the autoregressive decoding of LLMs without changing the output distribution. It shows notable speedup without a draft model and can linearly decrease the decoding steps with exponential investment in per-step FLOPs.

\nocite{langley00}
\bibliography{main}
\bibliographystyle{icml2024}

\newpage
\appendix
\onecolumn
\section{Algorithms}\label{app:verify}

\begin{algorithm}[!h]
   \caption{Jacobi decoding}
   \label{alg:jacobi}
\begin{algorithmic}[1]
   \STATE {\bfseries Input:} prompt $\vx^0$, model $p_M$, generation length $m$
   \STATE Initialize $\vy^0=(y_1^0,y_2^0,...,y_m^0)$
   \STATE Initialize $\vy^{output}\leftarrow()$
   \FOR{$i=1$ {\bfseries to} $m$}
   \STATE $\vy^{i}_{1:m}\leftarrow\arg\!\max(P_M(\vy^{i}_{1:m}|\vy^{i-1}_{1:m},\vx^0))$
   \STATE $\vo\leftarrow\vy^i$
   \STATE $stop\leftarrow$\textsc{StopCondition}$(\vy^{i},\vy^{i-1})$
   \IF{$stop$}
   \STATE break
   \ENDIF 

   \ENDFOR
   \STATE {\bfseries Output:} $\vo=(y_1,y_2,...,y_m)$
\end{algorithmic}
\end{algorithm}

\begin{algorithm}[!h]
   \caption{Lookahead decoding}
   \label{alg:lookahead}
\begin{algorithmic}[1]
   \STATE {\bfseries Input:} prompt $\vx^0=(x_1,x_2,...x_n)$, model $P_M$, n-gram size $N$, window size $W$, max \#speculations $G$, max steps $m$. 
   \STATE Initialize n-gram pool $\vC\leftarrow\emptyset$ 
   \STATE Initialize $\vo\leftarrow\emptyset$
   \STATE Randomly initialize 2D window $\vw^{2-N:0}_{1:W}$
   \STATE Set $\vo_{0}\leftarrow x_n$
   
   \FOR{$i=1$ {\bfseries to} $m$}
   \IF{$size(\vo)>=i$}
   \STATE Randomly set $\vw_{1:W}^i$
   \STATE continue
   \ENDIF
    \STATE \COMMENT{\textcolor{violet}{Lookahead Branch}}  
   \FOR{$j=1$ {\bfseries to} $W$}  
   \IF{$j=1$}
    \STATE $w^i_j\leftarrow\arg\!\max P_M(w_j^i|\vw^{i+2-N:i-1}_{j},\vo_{1:i},\vx^0)$ 
   \ELSE
   \STATE $w^i_j\leftarrow\arg\!\max P_M(w_j^i|\vw^{i+2-N:i-1}_{j},\vw_{2:j}^{i+1-N},\vo_{1:i},\vx^0)$
   \ENDIF
   \ENDFOR
   \STATE $\vw_{1:W}^i=(w_1^i,w_2^i,...,w_W^i)$
   \STATE \COMMENT{\textcolor{violet}{Verification Branch}}
   \STATE $\vg\leftarrow\emptyset$ 
   \FOR{$j=1$ {\bfseries to} $G$}
    \STATE $\vg^i\leftarrow$ n-gram from $\vC$ starting with $\vo_{i-1}$
   \ENDFOR
   \STATE \COMMENT{\textcolor{violet}{Verification Algorithm in Algo.~\ref{alg:greedy} and Algo.~\ref{alg:sample}.}}  
   \STATE $\vo$.append(\textsc{Verification$((x^0,\vo_{1:i}),P_M,\vg)$})
    \STATE \COMMENT{\textcolor{violet}{Update n-gram pool}}  
    \FOR{$j=1$ {\bfseries to} $W$}
    \STATE add n-gram $\vw^{i-N+1:i}_j$ to $\vC$
    \ENDFOR 
   \ENDFOR

    \STATE {\bfseries Output:} $\vo_{1:m}=(y_1,y_2,...,y_m)$

\end{algorithmic}
\end{algorithm}

\newpage

\begin{algorithm}[!h]
   \caption{Greedy Verification with \xxx}
   \label{alg:greedy}
\begin{algorithmic}[1]
\STATE {\bfseries input}  prefill $\vx^0$,  model $P_M$, n-grams $\vg^i$ with $i\in[1,G]$ 
\STATE {\bfseries output}  $\vo$ \COMMENT{\textcolor{violet}{accepted tokens of length 1 to N}}
\FUNCTION{GreedyVerification($\vx^0$, $P_M$, $\vg$)}
\STATE $\vV,\vD,\vo \leftarrow \emptyset,\emptyset,\emptyset$
\FOR{$i=1$ to $G$}
\STATE $\vV$.append($\vg^{i}_{2:}$) \COMMENT {\textcolor{violet}{each is a n-1 gram}}
\STATE $\vD$.append($P_M(\vg^{i'}_{2:},\vx_{next}|\vg^i_{2:},\vx^0)$) 
\STATE \COMMENT {\textcolor{violet}{obtain last token of $\vx^0$ and all $\vg^i_{2:}$'s outputs -- totally N probability distributions}}
\ENDFOR
\FOR{$i=1$ to $N-1$}
\STATE $j\leftarrow1$
\STATE $is\_accept\leftarrow0$
\STATE $\vP\leftarrow \vD$[$j$]$_i$ \COMMENT {\textcolor{violet}{$\vD$[$j$] is a series of N probability distributions;all $\vD[j]_{i}$ should be the same as different distributions are removed; size($\vD$)$>0$ is guaranteed}}
\WHILE{$j\leq$ size($\vV$)}
\STATE $s_j\leftarrow \vV$[$j$]$_i$ %
\IF{$s_j=\arg\!\max \vP$}
\STATE \COMMENT {\textcolor{violet}{accepted, update all potential speculations and probabilities}}
\STATE $\vo$.append($s_j$) 
\STATE $is\_accept\leftarrow1$
\STATE $\vV_{new},\vD_{new}\leftarrow\emptyset,\emptyset$
\FOR{$k=j$ to size($\vV$)}
\IF{$s_j$=$\vV$[$k$]$_{i}$}
\STATE $\vV_{new}$.append($\vV[k]$)
\STATE $\vD_{new}$.append($\vD[k]$)
\ENDIF
\ENDFOR 
\STATE $\vV,\vD\leftarrow \vV_{new},\vD_{new}$
\STATE break
\ELSE
\STATE \COMMENT {\textcolor{violet}{rejected, go to next speculation }}
\STATE $j \leftarrow j + 1$ 
\ENDIF
\ENDWHILE
\IF{$is\_accept$}
\STATE continue 
\ELSE 
\STATE \COMMENT {\textcolor{violet}{guarantee one step movement}}
\STATE $\vo$.append($\arg\!\max \mathcal{P}$)
\STATE break
\ENDIF 
\ENDFOR 
\IF{$is\_accept$}
\STATE $\vo$.append($\arg\!\max \vD[1]_N$)
\ENDIF 
\STATE {\bfseries return} $\vo$
\ENDFUNCTION

\end{algorithmic}
\end{algorithm}

\newpage

\begin{algorithm}[!h]
   \caption{Sample Verification with \xxx}
   \label{alg:sample}
\begin{algorithmic}[1]
\STATE {\bfseries input}  prefill $\vx^0$,  model $P_M$, n-grams $\vg^i$ with $i\in[1,G]$
\STATE {\bfseries output}  $\vo$ \COMMENT{\textcolor{violet}{accepted tokens of length 1 to N}}
\FUNCTION{SampleVerification($\vx^0$, $P_M$, $\vg$)}
\STATE $\vV,\vD,\vo \leftarrow \emptyset,\emptyset,\emptyset$
\FOR{$i=1$ to $G$}
\STATE $\vV$.append($\vg^i_{2:}$)\COMMENT {\textcolor{violet}{each is a n-1 gram}}
\STATE $\vD$.append($P_M(\vg_{2:}^{i'},\vx_{next}|\vg_{2:}^i,\vx^0)$) 
\STATE \COMMENT {\textcolor{violet}{obtain last token of $\vx^0$ and all $\vg^i_{2:}$'s outputs -- totally N probability distributions}}
\ENDFOR
\FOR{$i=1$ to $N-1$}
\STATE $j\leftarrow1$
\STATE $is\_accept\leftarrow0$
\STATE $\vP_j\leftarrow \vD$[$j$]$_i$ \COMMENT {\textcolor{violet}{$\vD$[$j$] is a series of N probability distributions;all $\vD$[$j$]$_{i}$ should be the same; size($\vD$)$>0$ is guaranteed}}
\WHILE{$j\leq$ size($\vV$)}
\STATE $s_j\leftarrow \vV$[$j$]$_i$ %
\STATE sample $r \sim U(0, 1)$
\IF{$r\leq \vP_j(s_j)$}
\STATE \COMMENT {\textcolor{violet}{accepted, update all potential speculations and probabilities}}
\STATE $\vo$.append($s_j$) 
\STATE $is\_accept\leftarrow1$
\STATE $\vV_{new},\vD_{new}\leftarrow\emptyset,\emptyset$
\FOR{$k=j$ to size($\vV$)}
\IF{$s_j$=$\vV$[$k$]$_{i}$}
\STATE $\vV_{new}$.append($\vV[k]$)
\STATE $\vD_{new}$.append($\vD[k]$)
\ENDIF
\ENDFOR 
\STATE $\vV,\vD\leftarrow \vV_{new},\vD_{new}$
\STATE break
\ELSE
\STATE \COMMENT {\textcolor{violet}{rejected, go to next speculation }}
\STATE $\vP_j(s_j)=0$
\STATE $\vP_{j+1}=\textrm{norm}(\vP_j)$
\STATE $j \leftarrow j + 1$ 
\ENDIF
\ENDWHILE
\IF{$is\_accept$}
\STATE continue 
\ELSE 
\STATE \COMMENT {\textcolor{violet}{guarantee one step movement}}
\STATE sample $\vx_{next}\sim \vP_j$
\STATE $\vo$.append($\vx_{next}$)
\STATE break
\ENDIF 
\ENDFOR
\IF {$is\_accept$}
\STATE $\vo$.append(sample $\vx_{next}\sim \vD[1]_N$)
\ENDIF 
 \STATE {\bfseries return} $\vo$
\ENDFUNCTION

\end{algorithmic}
\end{algorithm}

\section{Proof: Output distribution preserved disjoint n-gram verification}\label{app:prof-sample}
The sampling verification in \xxx is adapted from the algorithm in Specinfer but with all speculations generated by the greedy sample. It does not change the output distribution from a fundamental point that how the draft model generates speculations is unimportant.  

\textbf{Theorem A} For a given LLM, prompt and previously generated tokens $\vx=(x_1, x_2,...,x_{i})$, and $G$ speculations $\vs=(s_1,s_2,...,s_G)$ of next token $x_{i+1}$. Each speculation token is sampled by a greedy sample (i.e., probability of 1). We use $P(v|\vx)$ to represent the probability of $x_{i+1}=v$ sampled by the LLM and use $Q(v|\vx)$ to represent the probability of $x_{i+1}=v$ sampled by our proposed algorithm~\ref{alg:sample}. We use $P(v)$ and $Q(v)$ for short. We need to prove $P(v)=Q(v)$ for any $G$, and any $v$ and $s_j$ from the full vocabulary $V$. 
\begin{proof}
The proof of this part corresponds to line 14 to line 44 in algorithm~\ref{alg:sample}. Given speculations $\vs$, we use $a_j(v)$ to represent the probability that the token $v$ is accepted by the $j$-th speculation (line 18-line 30), and $r_j(s_j)$ is the probability that the token $s_j$ is rejected by the $j$-th speculation (line 30-line 35), where $s_j$ is the $j$-th speculation's token. Moreover, $a_{G+1}'(v)$ is the probability of being accepted by the sampling at line 41. For simplicity, we use $a_j$ to represent $a_j(v)$, $a_j'$ to represent $a_j'(v)$, and use $r_j$ to represent $r_j(s_j)$. We use $\vP_1$ to represent the probability distribution obtained in line 13 and $\vP_j$ to present the updated probability before the $j$-th speculation. We have $\vP_1(v)=P(v)$ as $\vP_1(v)$ is never updated. We define $Q_G(v)$ is the probability of $x_{i+1}=v$ sampled by algorithm~\ref{alg:sample} when we have $G$ speculations. Then we should have:

$Q_G(v)=a_1+r_1a_2+r_1r_2a_3+...+a_{G}\prod \limits_{k=1}^{G-1}r_k+a_{G+1}'\prod \limits_{k=1}^{G}r_k$

We use induction to prove $Q_G(v)=P(v)$ for any $G\geq1$, any $v\in V$, and any $s_j\in V$ with $1\leq j\leq G$:

\begin{enumerate}[leftmargin=15pt]
 \item [1)] When $G=1$, we have $Q_G(v)=a_1+r_1a_{2}'$. The initial guess $s_1$ can be either the same at $v$ or be different from $v$. 
\begin{enumerate}[leftmargin=15pt]
 \item [(1)] When $s_1=v$,  $a_1$ equals $\mathcal{P}_1(v)$ at line 17, which is the same as $P(v)$ as it is never updated. Upon this, we have $r_1=1 - a_1=1-\vP_1(v)=1-P(v)$. And, $a_{2}'$ is the updated probability $\vP_2(v)$ at line 42. Since $\vP_2(v)$ is set to zero once rejected at line 32, $a_{2}'=0$. In this case, $Q_G(v)=P(v)+(1-P(v))*0=P(v)$. 

 \item [(2)]  When $s_1\neq v$, $a_1$ should be $0$ even if $s_i$ is accepted. Moreover, we have $r_1=1-\vP_1(s_1)$. Then $\vP_2(v)$ is updated to $\frac{\vP_1(v)}{1-\vP_1(s_1)}$ at lines 32 and 33. Then $a_{2}’=\vP_2(v)$. In this case, $Q_G(v)=0+r_1*\frac{P(v)}{r_1}=P(v)$.
\end{enumerate}
    \item [2)] When $G=g$ holds, which means $Q_g(v)=a_1+r_1a_2+...+a_g\prod \limits_{k=1}^{g-1}r_k+a_{g+1}'\prod \limits_{k=1}^{g}r_k=P(v)$ for any $s_j,v\in V$, $1\leq j\leq g$.

We prove $Q_{g+1}(v)=Q_g(v)-a_{g+1}'\prod \limits_{k=1}^{g}r_k+a_{g+1}\prod \limits_{k=1}^{g}r_k+a_{g+2}'\prod \limits_{k=1}^{g+1}r_k=P(v)$ for the same $s_j,v\in V$, $1\leq j\leq g$, and any $s_{g+1}\in V$.

    \begin{enumerate}[leftmargin=15pt]
 \item [(1)]   When $s_g\neq v$, we have $a_g=0$. If $\vP_{g}(v)=0$, we have all $a_{g+1}'=0$,  $a_{g+1}=0$ and $a_{g+2}'=0$. It ensures that $Q_{g+1}(v)=Q_g(v)-0+0+0=Q_g(v)=P(v)$. 
 
 If $\vP_{g}(v)\neq 0$, $\vP_{g+1}(v)=\frac{\vP_{g}(v)}{r_g}=\frac{\vP_{1}(v)}{\prod \limits_{k=1}^{g}r_k}$ since $s_g\neq v$ by observation. 

Then we have $Q_{g+1}(v)=Q_g(v)-\vP_{g+1}(v)\prod \limits_{k=1}^{g}r_k+a_{g+1}(v)\prod \limits_{k=1}^{g}r_k+a_{g+2}'\prod \limits_{k=1}^{g+1}r_k=Q_g(v)-\frac{\vP_{1}(v)}{\prod \limits_{k=1}^{g}r_k}\prod \limits_{k=1}^{g}r_k+a_{g+1}\prod \limits_{k=1}^{g}r_k+a_{g+2}'\prod \limits_{k=1}^{g+1}r_k=Q_g(v)-P(v)+a_{g+1}(v)\prod \limits_{k=1}^{g}r_k+a_{g+2}'\prod \limits_{k=1}^{g+1}r_k$. Here we have another two cases:

\quad \ding{192} If $s_{g+1}=v$, $a_{g+1}\prod \limits_{k=1}^{g}r_k=\frac{\vP_1(v_i=v)}{\prod \limits_{k=1}^{g}r_k}\prod \limits_{k=1}^{g}r_k=P(v)$ and $a_{g+2}'=\vP_{g+2}(v)=0$. We have $Q_{g+1}(v)=Q_g(v)-P(v)+P(v)+0=Q_g(v)=P(v)$.

\quad \ding{193} If $s_{g+1}\neq v$, $a_{g+1}=0$. $a_{g+2}'\prod \limits_{k=1}^{g+1}r_k=\vP_{g+2}(v)\prod \limits_{k=1}^{g+1}r_k=\frac{\vP_{1}(v)}{\prod \limits_{k=1}^{g+1}r_k}\prod \limits_{k=1}^{g+1}r_k= P(v)$. So $Q_{g+1}(v)=Q_g(v)-P(v)+0+P(v)=Q_{g}(v)=P(v)$

 \item [(2)] When $s_g=v$, $\vP_{g+1}(v)$ is set to zero at line 32 after this step. In this case, we have $a_{g+1}'=\vP_{g+1}(v)=0$,  $a_{g+1}=\vP_{g+1}(v)=0$ and $a_{g+2}'=0$. It makes that $Q_{g+1}(v)=Q_g(v)-0+0+0=Q_g(v)=P(v)$

\end{enumerate}
\end{enumerate}

\end{proof}

This part of the proof guarantees that from line 14 to line 44 in Algorithm~\ref{alg:sample}, any new token appended to $\vo$ can follow the original distribution of the LLM. Line 21 to line 28 guarantees that sequences in $\vV$ share the same prefix of length $i-1$ in every iteration. This further guarantees that $\vP$ from $\vD[j]_i$ is the same for all $j$, follows the wanted distribution. Thus, the correctness of the whole sampling algorithm is proved.

\section{Derivation of Expectation of The Number of Accepted Tokens}\label{app:exp}
We first start with single-candidate speculation. We need to obtain the probability of accepting $i$ tokens as $P(\#accepted\ tokens=i)$ for all possible $i$. Since the speculation's length is $\gamma$, the probability of accepting $i$ tokens with $i\geq\gamma+2$ is 0. $P(\#accepted\ tokens=1)$ is the probability of the first token being rejected, which is $1-\alpha$. The probability $P(\#accepted\ tokens=i)=P(\#accepted\ tokens=i-1)*\alpha$, for all $i\leq\gamma$. The probability $P(\#accepted\ tokens=\gamma+1)$ is accepting all tokens, which is $\alpha^{\gamma}$.  Thus we have the following, which is Eq.~\ref{for:spec}:

\begin{eqnarray}
\footnotesize
\label{dev:spec}
 E(\#tokens) &=& \sum\limits_{i=1}^{\gamma+1}i*P(\#accepted\ tokens=i) \nonumber\\
 & =& 1  * (1-\alpha)  + 2 * (1 - \alpha) * \alpha + ...\nonumber  + (\gamma+1) * \alpha^\gamma\nonumber\\
 & =& (1-\alpha) + (2\alpha-2\alpha^2)+(3\alpha^2-3\alpha^3)+...+ (\gamma+1)\alpha^{\gamma} \nonumber\\ 
 & =& 1 + (-\alpha+2\alpha)+(-2\alpha^2+3\alpha^2)+(-3\alpha^3+4\alpha^3)+...+(\gamma+1)\alpha^{\gamma} \nonumber\\
 & =& 1 + \alpha + \alpha ^ 2 + \alpha ^3 + ... + \alpha ^ \gamma\nonumber \\
 & =& \frac{1-\alpha^{\gamma + 1}}{1-\alpha}       
\end{eqnarray}

We then investigate the case of speculations with a batch size of $b$. We need to obtain the probability of accepting $i$ tokens as $P(\#accepted\ tokens=i)$. Since all speculations' length is $\gamma$, the probability of accepting $a$ tokens with $a\geq\gamma+2$ is 0. We use $p_i$ to denote $(1-\alpha^i)^b$, which is the probability that at most $i$ tokens are accepted in all $b$ speculations. For all $i\leq\gamma$, we should have $P(\#accepted\ tokens=i)=p_i-p_{i-1}$. And, the probability $P(\#accepted\ tokens=\gamma+1)$ should be $(1-p_{\gamma})$.  Thus we have the following, which is Eq.~\ref{for:bspec}:

\begin{eqnarray}
\footnotesize
\label{dev:bspec}  
E(\#tokens) &=& \sum\limits_{i=1}^{\gamma+1}i*P(\#accepted\ tokens=i) \nonumber\\
& =& \sum\limits_{i=1}^{\gamma}{i(p_i-p_{i-1})} \nonumber + (\gamma+1)*(1-p_{\gamma}) \nonumber\\
&=& (p_1-p_0)+(2p_2-2p_1)+(3p_3-3p_2)+...+(\gamma+1)(1-p_\gamma) \nonumber\\ 
&=& -p_0+(p_1-2p_1)+(2p_2-3p_2)+(3p_3-4p_3)...+(\gamma+1) \nonumber\\
&=& -p_0-p_1-p_2-...-p_{\gamma}+(\gamma+1) \nonumber \\
& = &(\gamma+1)-\sum\limits_{i=1}^{\gamma}(1-\alpha^i)^b    
\end{eqnarray}

\newpage
\section{Prompt for LLaMA-2-Chat on Summarization Tasks}\label{app:sum}
We use the following as the prompt for summarization task, modified from~\cite{aws-prompt}.

\begin{center}
\begin{tabular}{l}
\toprule
\ding{228} Prompt:\\
$[$INST$]$ $<<$SYS$>>$ \\

You are an intelligent chatbot. Answer the questions only using the following context:\\

\{Original Text\}\\ 

Here are some rules you always follow:\\

- Generate human readable output, avoid creating output with gibberish text.\\

- Generate only the requested output, don't include any other language before or after the requested output.\\

- Never say thank you, that you are happy to help, that you are an AI agent, etc. Just answer directly.\\

- Generate professional language typically used in business documents in North America.\\

- Never generate offensive or foul language.\\

$<</$SYS$>>$\\

Briefly summarize the given context. $[/$INST$]$\\

Summary: \\
\bottomrule
\end{tabular}
\end{center}

\section{Verification of Generation Quality for Greedy Sampling and Advanced Supports}\label{app:verify}
\textbf{Generation Quality with Greedy Search is not changed.} 
Theoretically, \xxx does not change the output generation of greedy search due to the verification mechanism. However, \xxx's output does not perfectly align with the huggingface's implementation of greedy search in practice. We attribute this discrepancy to numerical accuracy issues. To substantiate this claim, we compared the output results as follows. We use the LLaMA-2-7b-Chat model's single precision (FP32) inference with huggingface's greedy search on 160 turns on the MT-Bench dataset as a baseline. With single precision inference, the outputs of \xxx (on 1GPU, 4GPUs, and 8GPUs) are the same as the output of the baseline. With half-precision (FP16) inference, huggingface's greedy search has 35 out of 160 (w/o FlashAttention) and 42 out of 160 (w/ FlashAttention) answers not perfectly aligned with the baseline output. In contrast, \xxx and its integration with FlashAttention and multi-GPU inference has 35-44 results different from the baseline output under different settings. We claim this result can show that \xxx can retain the output distribution using a greedy search within the numerical error range (not worse than huggingface's half-precision inference). Besides, Tab.~\ref{tab:sample-perf} also strengthens the statements for greedy search.

\textbf{Generation Quality with LP and FlashAttention Augmentation is not changed.} 
We verify that FlashAttention and LP Support will not change the compression ratio ($\vS$) of vanilla \xxx. We compared each 18 generations of \xxx w/ FlashAttention and w/o FlashAttention (7B and 13B model on MT-Bench, HumanEval, and ClassEval); the average $\vS$ w/ FlashAttention is 3.267  while w/o FlashAttention is 3.259, with less than 0.3\% differences. We also compared 6 generations of \xxx on a single GPU and 12 generations with LP (7B model on MT-Bench, HumanEval, and ClassEval, both with $N=5$, $W=15$, and $G=15$). The average $\vS$ on a single GPU is 2.558, while on multiple GPUs, it is 2.557, with less than 0.1\% differences. We claim that our advanced support does not change $\vS$.

\end{document}